\documentclass[journal,twoside,web]{ieeecolor2}
\usepackage{generic}
\usepackage{amsmath,amssymb,amsfonts}
\usepackage{hyperref}
\usepackage{algorithmic}
\usepackage{graphicx}
\usepackage{textcomp}
\usepackage{xcolor}

\begin{document}

\title{Towards Investigating Residual Hearing Loss: Quantification of Fibrosis in a Novel \newline Cochlear OCT Dataset}

\author{Julia Dietlmeier, \IEEEmembership{Member, IEEE}, Benjamin Greenberg, \IEEEmembership{Graduate Student Member, IEEE}, Wenxuan He, Teresa Wilson, Rubing Xing, Jordan Hill, Adrienne Fettig, Madeline Otto, Teyhana Rounsavill, \newline Lina A. J. Reiss, Jingang Yi, \IEEEmembership{Senior Member, IEEE}, Noel E. O'Connor, \IEEEmembership{Member, IEEE}, and \newline George W.S. Burwood
\thanks{This work was supported by funding from Hearing Health
Foundation Emerging Research Grant 969672 (Burwood).
Research reported in this publication was supported by the
National Institute On Deafness And Other Communication
Disorders of the National Institutes of Health under Award
Number R21DC020794 (Burwood).}
\thanks{J. Dietlmeier and N. E. O'Connor are with the Insight Research Ireland Centre for Data Analytics, Dublin City University, Ireland (e-mail: julia.dietlmeier@insight-centre.org, noel.oconnor@insight-centre.org). }
\thanks{B. Greenberg and J. Yi are with the Rutgers University, New Jersey, USA (e-mail: brg60@scarletmail.rutgers.edu, jgyi@soe.rutgers.edu). }
\thanks{W. He, T. Wilson, R. Xing, J. Hill, A. Fettig, M. Otto, T. Rounsavill, L. A. J. Reiss, G. W. S. Burwood are with the Oregon Health \& Science University (OHSU), Oregon Hearing Research Center (OHRC), Portland, Oregon, USA (e-mail: burwood@ohsu.edu).}}

\maketitle

\begin{abstract}
Objective:
Cochlear implants (CIs) are bionic prostheses that restores hearing via electrical stimulation of the auditory nerve. Hybrid CIs, which use electroacoustic stimulation (EAS), combine residual low-frequency acoustic hearing with CI electrical stimulation. Intracochlear fibrosis, which forms in response to the presence of the implant, may impede residual hearing function and gradually reduce the efficacy of EAS. It is therefore a translational objective to study the formation of cochlear fibrosis in rodents, with the goal of reducing fibrotic burden and improving outcomes for CI patients. 

Methods:
We generate and annotate a novel dataset of optical coherence tomography (OCT) images from chronically implanted guinea pigs as part of an ongoing study focused on implant induced fibrosis. Objectively assessing fibrotic burden in this model, with high resolution and repeatability, presents an obvious use case for computer vision methods. 

Results:
We present the results of several state-of-the-art semantic segmentation models and compare their efficacy for identifying cochlear fibrosis and other relevant annotations, using a new library of manually segmented OCT images. 

Conclusions:
We find that the best performance is achieved by using a modified version of the well-known UNET architecture (which we term 2D-OCT-UNET) that operates on the upscaled OCT input resolution.

Significance:
For the first time, we have successfully applied computer vision techniques to an OCT dataset of implanted cochleae with fibrosis. Using this deep learning model, the cochlear fibrotic burden calculation can be reliably carried out as we verify in our experimental section. The dataset and the project code are available at: \href{https://github.com/juliadietlmeier/CF-OCT-segmentation}{https://github.com/juliadietlmeier/CF-OCT-segmentation}. 

\end{abstract}

\begin{IEEEkeywords}
Cochlear fibrosis, cochlear implants, deep learning, residual hearing loss, semantic segmentation
\end{IEEEkeywords}

\section{Background}\label{}
Cochlear implants (CIs) are a major success story in the field of bionic prostheses. Study of their development, function, and interaction with human physiology is of continued interest to the field of biomedical engineering e.g. \cite{kipping2024computational}. The CI is comprised of a flexible array of electrodes inserted into the scala tympani (ST) of the cochlea, and acoustic signals from an over-ear receiver are transduced into pulse trains delivered to an individual electrode depending upon the acoustic frequency. CIs restore some hearing in deaf patients. Others with residual low frequency hearing can receive so-called electroacoustic stimulation (EAS), that combines amplification of low frequency acoustic signals from a hearing aid with the electrical stimulation of the cochlea at high frequencies \cite{gantz2003combining}. EAS patients often demonstrate improved speech-in-noise intelligibility and musicality when compared to conventional CI patients \cite{gantz2005preservation}, \cite{turner2004speech}. 

Unfortunately, up to 50\% of EAS patients lose their residual hearing in months to years following CI \cite{gantz2009hybrid}, \cite{quesnel2016delayed}. The causes for this residual hearing loss are unknown and multifactorial. Causes may include spiral ganglion neuronal loss, hair cell loss, surgical trauma, and immune response \cite{tanaka2014factors}, \cite{o2013relations}, \cite{eshraghi2013molecular}, \cite{choong2020nanomechanical}.

\begin{figure*}[h]
	\centering
		\includegraphics[width=17cm]{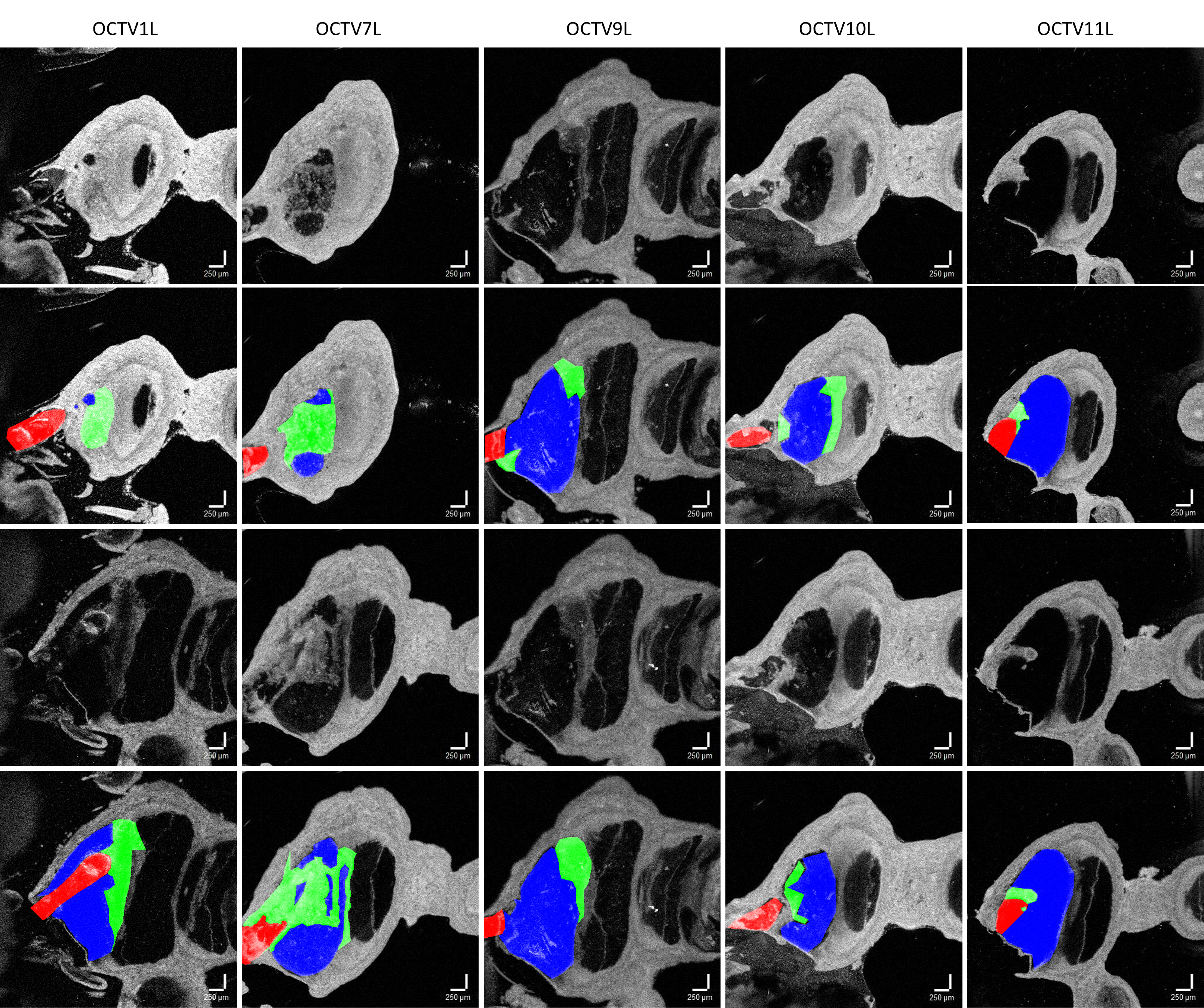}
	  \caption{Cochlear OCT dataset samples and the corresponding ground truth annotations from five annotated volumes. The CI/Track class is depicted in red, the Fibrosis class in green and the  ST/Free Space class in blue.}\label{fig:dataset_samples}
\end{figure*}

The mechanical function of the mammalian cochlea, whereby a base-to-apex travelling wave is induced by pressure differences across the mechanosensitive organ of Corti \cite{robles2001mechanics}, is hypothesised to be disrupted by inflammatory tissue response, which includes the formation of fibrotic tissue and eventual ossification of the CI array e.g. \cite{tanaka2014factors},  \cite{noonan2020immune}, \cite{simoni2020immune}. Modelling \cite{choi2005predicting}, and acute rodent studies \cite{lee2024guinea} have suggested that CI and/or fibrotic burden could interfere with the mechanical function of the cochlea. A recent report analyzing multiple histological and physiological indices of residual hearing loss in chronically implanted rodents could not exclude the contribution of fibrosis-induced conductive hearing loss \cite{reiss2024chronic}. 

The study of fibrotic formation, its effects upon cochlear mechanics, and whether it is a cause of residual hearing loss is ongoing, and the adoption of Optical Coherence Tomography (OCT) in CI research is recent \cite{hyzer2024effects}. Novel imaging strategies such as OCT have the potential to produce large datasets. These datasets present a use case for machine learning driven segmentation algorithms in order to objectively and rapidly produce estimates of cochlear fibrotic burden with high resolution. We present the performance of several state-of-the-art CNN and Transformer-based image segmentation architectures and conclude that our approach, 2D-OCT-UNET, is the first successful application of computer vision techniques to a novel OCT dataset containing CI-induced intracochlear fibrosis. This tool can be used to assess multiple imaging modalities with low-resolution ground truth segmentation as an input.

\section{Novel Cochlear OCT dataset}\label{}
\begin{figure*}[t!]
	\centering
		\includegraphics[width=15cm]{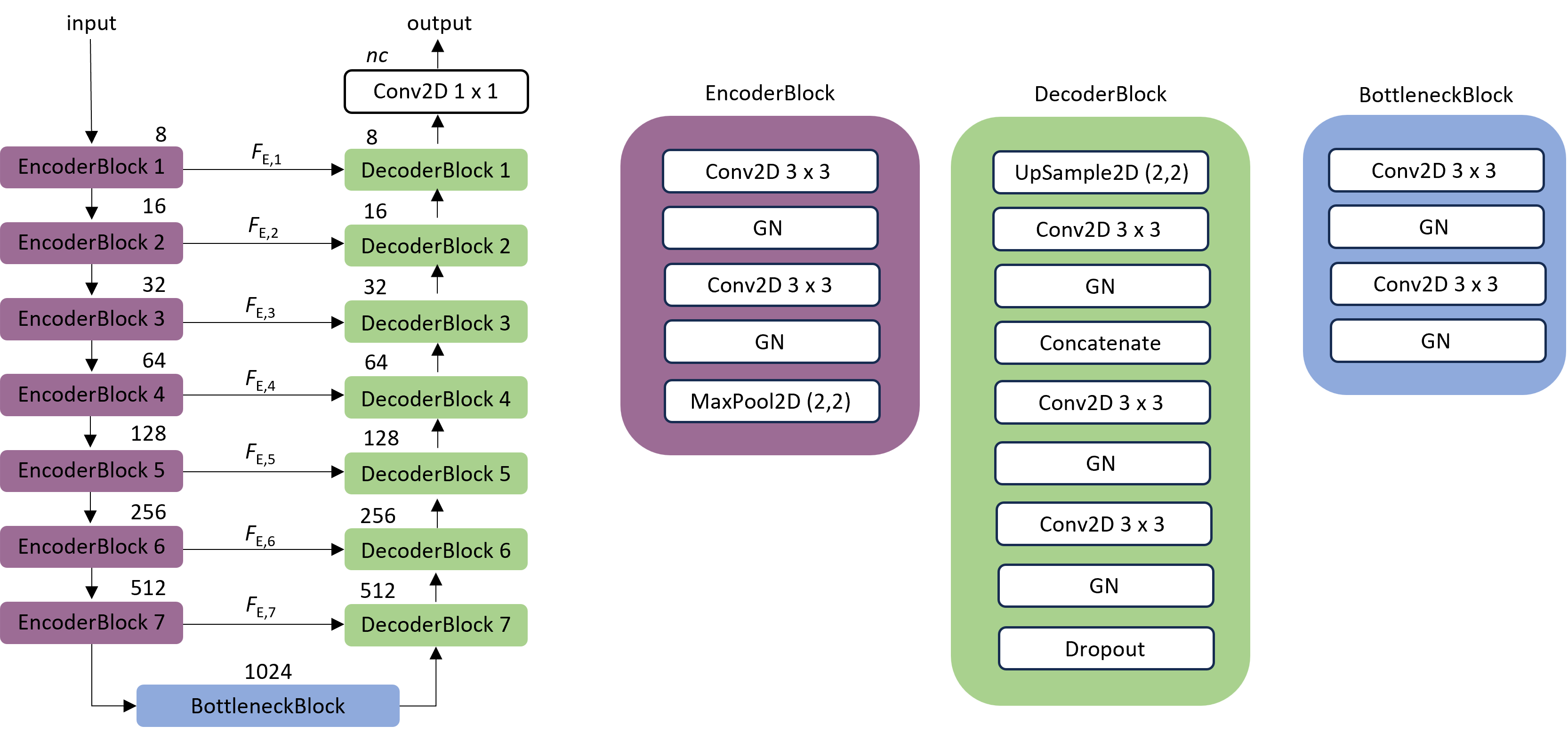}
	  \caption{Block diagram of the proposed 2D-OCT-UNET architecture for the multiclass OCT segmentation used in this work. The very deep 2D-OCT-UNET processes two-dimensional OCT slices and consists of seven encoder-decoder blocks with skip connections. The numbers above the encoder, bottleneck and decoder blocks indicate the number of filters in the convolutional \textbf{Conv2D} layers. GN stands for Group Normalization layers with the number of groups parameter $ng = 2$. The input resolution of the UNET is set to be 1024 $\times$ 1024 pixels. We include \textbf{Dropout}(0.1) layers only in the decoder. The number of filters in the last \textbf{Conv2D} layer is equal to the $nc=4$ (number of classes). }\label{fig:our_unet_architecture}
\end{figure*}

As part of a study exploring the relationship between cochlear fibrosis and residual hearing mechanics, we curate an annotated dataset of cochlear OCT images from five animals (volumes): OCTV1L, OCTV7L, OCTV9L, OCTV10L and OCTV11L. While other OCT datasets of middle and inner ear anatomy exist e.g. \cite{buswinka2024large}, \cite {kerkhofs2022optical}, \cite {liu2024dresden}, this dataset is the first of its kind to capture CI-induced intracochlear fibrosis in a chronically implanted rodent model using OCT. OCT is a label free, non-invasive technique for cross-sectional high-resolution tissue imaging, using low coherence infrared light interferometry, which provides images without histological sectioning of the tissue. The imaged volumes in this study are OCT B-scans that have been Z resliced to provide En Face images of the left (implanted) cochlea. Each OCT volume contains 1024 slices in total. 

We annotate only a small number of slices from each volume with three distinct semantic classes to distinguish between fibrosis, ST and CI areas of the cochlea. The total number of annotated images amounts to 173. Selected dataset samples and the corresponding annotations are provided in Fig.~\ref{fig:dataset_samples}. The dataset is challenging because of the high variability of shapes, sizes and appearances of the three objects of interest. All three classes partially cover ``empty'' areas (black pixels), and fibrotic tissue closely abutts normal tissue. 

Annotations were performed by experienced segmenters (Oregon Health \& Science University School of Medicine) using Hasty.ai manual segmenation tools\footnote{https://hasty.cloudfactory.com/}. Each image is annotated with three object classes: ``ST/Free Space'' (label `1'), ``CI/Track'' (label `2') and ``Fibrosis'' (label `3'). For consistency, we ensure that each image in the dataset has three annotated objects. We further split the dataset into 80\% train and 20\% test subsets using the stratified random sampling approach in order to preserve the original dataset distribution. The dataset is therefore imbalanced (from the classification perspective) as OCTV1L images have the largest proportion of 77 slices in the resulting cohort of 173 annotated slices. The resulting training subset contains 139 images and the testing subset contains 34 images. The dataset is also highly imbalanced from the perspective of semantic segmentation as can be observed e.g. from Fig. \ref{fig:dataset_samples} where the three classes of interest constitute varying percentage of pixels and in some extreme cases this percentage is much less than the number of pixels allocated to the ``Background'' class. The original OCT image resolution is 400 $\times$ 400 pixels. To our best knowledge, this is the first-of-its-kind fully annotated OCT dataset for cochlear fibrosis studies. 

\section{Architectures for semantic segmentation of Cochlear OCT data}\label{}

Artificial intelligence (AI) is revolutionizing computer vision but the application of AI to mass processing of medical images in practice remains quite novel. From a computer vision perspective, AI is a set of artificial deep neural network architectures such as Convolutional Neural Networks (CNNs) and Transformers. CNNs were developed first in 1998 for handwritten digit recognition \cite{LeCun_1998} and have at least three types of hidden layers such as convolution, activation and pooling.  Convolution is the central operation to CNNs and consists of a dot product of a learnable kernel and a partial receptive field. CNNs are used for image classification, retrieval, reconstruction, detection and image segmentation computer vision tasks.

Transformer-based architectures are also deep artificial neural networks that were first applied to Natural Language Processing (NLP). Vision Transformer architectures (ViTs) emerged as a powerful alternative to CNNs and have been shown to outperform the latter on many downstream tasks if pretrained on large amounts of data. In contrast to CNNs, ViTs decompose the image into patches and process the information with the attention mechanism \cite{ViTs}.

A relatively new computer vision concept is the \textit{foundation} model. These are mostly vision Transformer-based models that are pretrained on a vast amount of images and that can be adapted (fine-tuned) for different tasks.

The main objective of semantic segmentation is to categorize each pixel in an image into an object. We have a multiclass (“Fibrosis”, “ST/Free Space”, “CI/Track” and “Background”) setting in this work where we endeavour to segment three distinct non-overlapping objects (plus background). For this task, we explore different semantic segmentation architectures and specifically investigate a UNET-based family of CNNs, a recent foundation model termed Segment Anything Model (SAM), a vision Transformer architecture termed SegFormer, and a recent semantic segmentation model termed MST-DeepLabv3+. We leverage transfer learning for two of the UNET-based models by using pretrained backbones. In the following sections, we describe these architectures in detail.

\subsection{Proposed 2D-OCT-UNET configuration}

Our processing pipeline is shown in Fig.~\ref{fig:our_unet_architecture}. We first modify the well-known fully-supervised UNET architecture \cite{unet}. UNET is a fully-convolutional deep neural network with a symmetric encoder-decoder structure and skip connections which allow the sharing of information at different scales. UNET is a classical image segmentation model which has also found its application in image reconstruction, some Transformer architectures (e.g. UNETr) \cite{UNETr} and even in Denoising Diffusion Probabilistic Models (DDPM) \cite{DDPM}. A well-known feature of UNET is that it can work well with little training data which is appropriate in our current experimental setting. 

UNET takes on an image as an input and first gradually extracts context features through a set of encoder blocks. The encoder blocks follow the typical architecture of a CNN. During this process, each feature map size is reduced by 2 and the number of feature channels is doubled. The decoder blocks gradually upsample, convolve and halve the number of feature channels. The final convolutional layer uses $1 \times 1$ convolution to map each feature vector to the number of classes. The purpose of skip connections is to recover the high-resolution information by alleviating the loss of spatial information in the encoder blocks \cite{skips}.

Our 2D-OCT-UNET architecture consists of seven encoder and decoder blocks and Group Normalization (\textbf{GN}) layers instead of more familiar Batch Normalization layers as illustrated in Fig. \ref{fig:our_unet_architecture}. Group Normalization \cite{GN_layers} has been shown to improve model performance for small batch sizes as in our case (batch size = 2). We increase the number of convolutional filters by a factor of two after every \textbf{MaxPool2D} layer and start with $n=8$ number of filters in the first layer. 

There are 37 convolutional \textbf{Conv2D} layers (with the kernel size of $3 \times 3$) in total. Increasing the number of layers (depth of the network) up to this configuration allows us to handle long-range feature dependencies \cite{TransUNET} - a mechanism usually attributed to the \textit{self-attention} property of Transfomer-based architectures. 

The 2D-OCT-UNET shown in Fig.~\ref{fig:our_unet_architecture} takes on pseudo-RGB (3 channels) inputs of size 1024 $\times$ 1024. Therefore, we resize all original 400 $\times$ 400 OCT images to this resolution. In this configuration, we train the 2D-OCT-UNET from scratch and do not use transfer learning. 

The \textbf{Conv2D} layer weights of the 2D-OCT-UNET are initialized with He Normal \cite{he_normal} initializer that draws samples from a truncated normal distribution centered on 0 with $\sigma = \sqrt(2 / f_{in})$ where $f_{in}$ is the number of input units in the weight tensor. The output of the model is in the form of $nc=4$ likelihood maps. The pixel-wise \textit{argmax} is taken to obtain the final segmentation result.

\subsection{VGG16-UNET}

The VGG16-UNET model \cite{vgg16_unet} used in this study incorporates a contracting path and an expanding path. The contracting path (encoder) is the VGG16 model \cite{vgg16} with the three fully connected layers removed. The architecture of the expanding path (decoder) resembles the upsampling path of the original UNET. In total, the VGG16-UNET has four encoder blocks, bottleneck and four decoder blocks which are connected to the VGG16-based encoder blocks via skip connections. The VGG16 encoder is pretrained on the ImageNet \cite{imagenet} dataset. The structure of a decoder block is as follows: \textbf{Conv2DTranspose}(kernel\_size=2, strides=2) $\rightarrow$ \textbf{Concatenate} $\rightarrow$ \textbf{conv\_block}. The structure of a \textbf{conv\_block} is \textbf{Conv2D}(kernel\_size=3) $\rightarrow$ \textbf{BatchNormalization} $\rightarrow$ \textbf{ReLU}. The number of convolutional filters decreases with the network depth such that the first decoder block has 512 filters, the second has 256 filters, the third has 128 filters and the fourth has 64 filters. We use 256 $\times$ 256 input image resolution.

\subsection{UEfficientNet}
\begin{figure}[h]
	\centering
		\includegraphics[width=5cm]{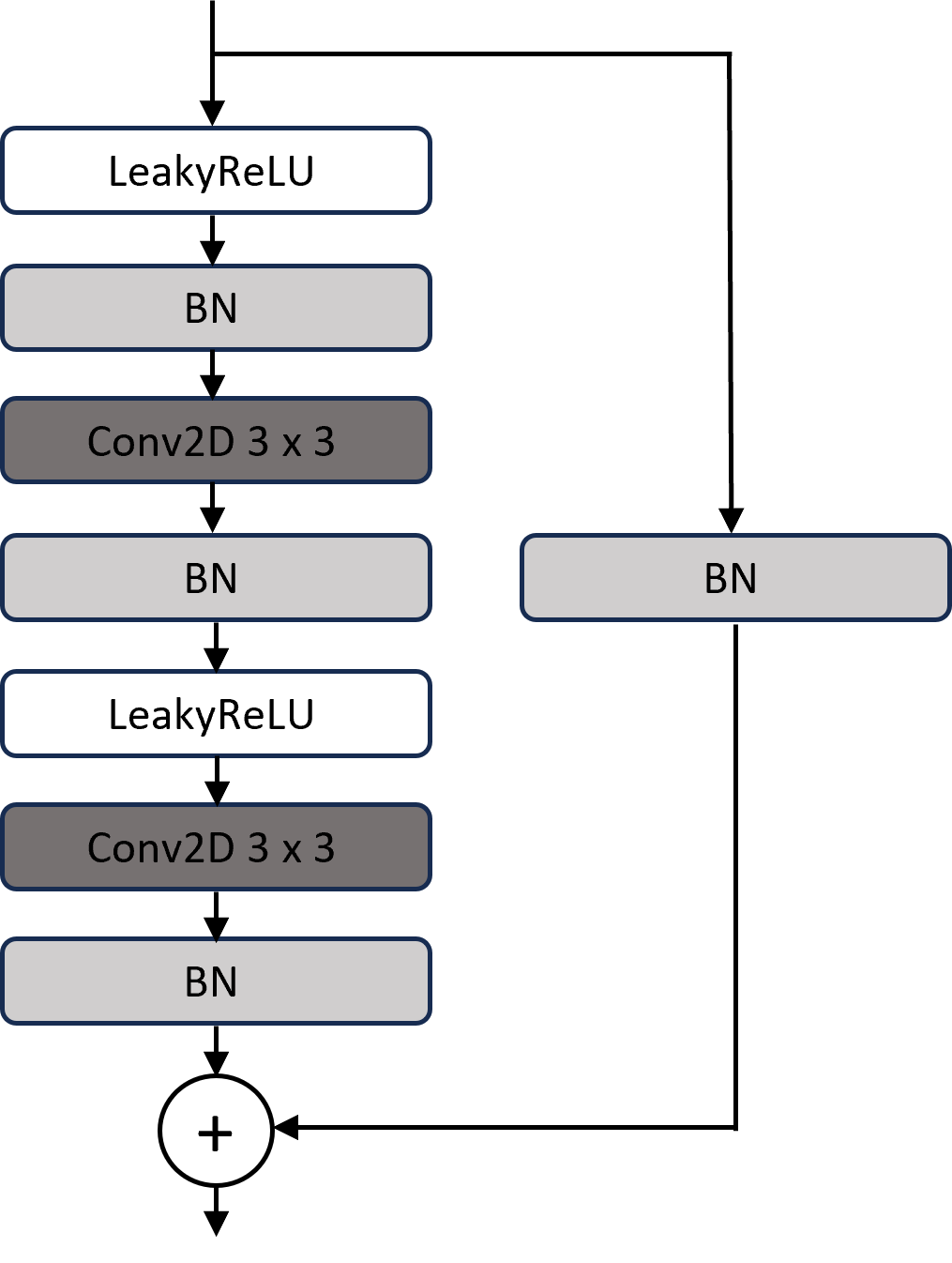}
	  \caption{Schematic of the residual unit in the decoder block of the UEfficientNet model.}\label{fig:residual_block}
\end{figure}

The UEfficientNet model \cite{UEfficientNet} is a combination of the EfficientNet encoder with UNET. Specifically, we used the EfficientNet-B4 \cite{efficientnet} encoder pretrained on the ImageNet dataset. The idea of the UEfficientNet follows the architecture of the VGG16-UNET with a difference in the design of the decoder blocks which employ residual learning (Fig. \ref{fig:residual_block}). 

EfficientNet uses a ``compound coefficient optimization'' technique to balance depth, breadth, and resolution scaling. This technique is one of the main features of the EfficientNet. This enables higher accuracy levels while making better use of computer resources \cite{efficientnet}. We use 256 $\times$ 256 input image resolution with the UEfficientNet model.
 
\subsection{SegFormer}

SegFormer is an efficient semantic segmentation enco-der-decoder architecture which unifies Transformers with lightweight multilayer perceptron (MLP) decoders in contrast to most decoder architectures using deconvolutions and upsampling operations. These decoders combine information from different layers and leverage both local and global attention to extract powerful representations. The hierarchically structured encoder outputs multiscale features and makes use of self-attention - a key mechanism for focusing on different parts of a sequence. A known issue is that Transformer models show decreased performance when the testing resolution differs from the training resolution. Specifically, the SegFormer model does not employ positional encoding and thus avoids the interpolation of positional codes \cite{segformer} yielding generally better performances. We load a pretrained hierarchical SegFormer model variant \texttt{nvidia/mit-b0} from Hugging Face Transformers\footnote{https://huggingface.co/nvidia/mit-b0}. We use the resolution of 512 $\times$ 512 pixels as an input to the SegFormer model. 

\subsection{MST-DeepLabv3+}

MST-DeepLabv3+ is the most recent semantic segmentation model presented in \cite{MST_DeepLab_nature2024}. It is based on the DeepLabv3+ \cite{DeepLab_2018} architecture and can produce better results with fewer training parameters due to the use of a lightweight convolutional backbone MobileNetV2. In addition this model leverages a channel attention mechanism called SENet (Squeeze-and-Excitation) \cite{SENet_2018} as well as transfer learning by using pretraining on the ImageNet dataset. We use the resolution of 256 $\times$ 256 pixels as an input to the MST-DeepLabv3+ model and the categorical crossentropy as a loss function.

\subsection{Segment Anything Model (SAM)}
The Segment Anything Model \cite{kirillov2023segany}, or SAM, is a recent image segmentation foundation model. Foundation models are starting to reshape computer vision. These are large pre-trained deep neural networks that can be leveraged for various downstream tasks in computer vision without training from scratch. Specifically, SAM is based on vision Transformers and trained on a dataset of over 11 million natural images and one billion segmentation masks \cite{kirillov2023segany}. It consists of three components: a pretrained Vision Transformer (ViT) image encoder, a prompt encoder that supports sparse and dense prompts (such as points, boxes and masks), and a mask decoder based on a modified Transformer decoder block. 

Though SAM has achieved strong zero-shot transfer, in order to perform in our prompt-free, multiclass application, we fine-tune the model using our new cochlear OCT dataset. This is to ensure strong performance for a fair comparison. Specifically, we leverage an adapter-based prompt-free fine-tuning approach developed for medical image segmentation \cite{gu2024build}. The ViT-B image encoding architecture is used and adapter blocks were added to the mask decoder.

\section{Methodology}\label{}
\subsection{CI animal model}
 
Imaging volumes were obtained from 5 male Hartley guinea pigs (Charles River Laboratories). Animals were anesthetized with intramuscular Ketamine (60 mg/kg) and Xylazine (5 mg/kg). Lidocaine (2mg/kg) was injected around the incision site for analgesia, and anesthesia was maintained using 2-3\% isoflurane in 100\% oxygen, at 3 l/min. Animals received CIs (HL08, Cochlear Ltd, Australia) at 10 weeks of age, and had a post-surgery latency period of 8-10 weeks to allow for the formation of intracochlear fibrosis. The implantation procedure has been previously described (\cite{tanaka2014factors} and \cite{hyzer2024effects}). Electrical stimulation of the implants, audiometric testing, optical coherence vibrometry testing and histology were conducted but are outside of the scope of this report. 

\subsection{Tissue preparation}
Following vibrometry experiments, animals were euthanized with an overdose of anesthetic. Further post-mortem vibrometry measurements were made, and then the animals were decapitated, and both temporal bones harvested, taking care not to strain the subcutaneous leads of CIs. The walls of the temporal bone were removed using a rongeur to expose the whole cochlea. A hole was made in the apex of the otic bone and the cochleae were slowly perfused with 4\% paraformaldehyde, before overnight fixation at 4\textdegree C. Temporal bones were then decalcified for up to two weeks in EDTA at 4\textdegree C. The EDTA was freshened daily. 

\subsection{OCT imaging}
Imaging was conducted using a Telesto III Spectral Domain OCT microscope (Thorlabs GmbH, Germany), central wavelength 1300 nm, through a 5x objective lens (Thorlabs LSM03). Images were acquired using ThorImage 4.4 (Thorlabs GmbH, Germany). Samples were placed in a 60 mm dish, unsubmerged, and the basal turn of the cochlea was placed in the coherence gate in 2D mode, and a $4 mm^2  \times 3.5 mm$, $400\times400\times1024$ voxel volume was acquired in 3D mode at a 10 kHz sampling rate. All three scalae of the basal turn could be imaged, including the round window and the CI insertion point. Cochleae were imaged with the CI in place, and then the array was carefully removed with tweezers while submerged in phosphate buffered saline, to reduce the chance of air ingress into the sample. Explanted cochleae were reimaged, and unimplanted right cochleae were also imaged. This dataset consists of explanted imaging volumes from the implanted cochleae only. 

Imaging spectra were extracted from .OCT file format and converted to 8-bit grayscale image sequences in ImageJ. Image sequences were uploaded to Hasty.ai for manual segmentation. 

\subsection{Implementation details}

All semantic segmentation models (except SAM) were implemented in Python 3.9.12, Tensorflow 2.9.1 and Keras 2.9.0. The SAM model was implemented in Python 3.9.12 and PyTorch 2.0.1 (cu117).

All experiments were performed on a desktop computer with the Ubuntu operating system 18.04.3 LTS with the Intel(R) Core(TM) i9-9900K CPU, Nvidia GeForce RTX 2080 Ti GPU, and a total of 62GB RAM. 

We train the 2D-OCT-UNET, UEfficientNet, VGG16-UNET and MST-DeepLabv3+ models using an Adam optimizer \cite{adam} with a learning rate of lr = 0.0001. Batch size is 2 and the models are trained for 100 epochs to ensure convergence. The best model checkpoint is used for inference. We use the region-based (geometric) multiclass Dice loss as the main optimization objective for the the 2D-OCT-UNET, UEfficientNet and VGG16-UNET models and the categorical crossentropy loss function for the MST-DeepLabv3+ model. 

Dice loss equals to 1 - DSC where DSC is the Dice Score Coefficient used to measure the overlap between the ground truth and the predicted segmentation. DSC ranges from 0 to 1 with 0 indicating no overlap and 1 indicating perfect overlap. It is empirically argued that a Dice loss is more robust to the extreme class imbalance than a Cross-Entropy loss \cite{Dice_imbalance}, \cite{Dice_imbalance_2}. Specifically a multiclass version of a Dice loss function has the following form \cite{dice_loss}: 
\begin{equation}
    Dice Loss = \frac{1}{C}\sum_{i=1}^C\biggl(1-\frac{2\sum_n A_{in}B_{in}+\gamma}{\sum_n A_{in}^2+\sum_n B_{in}^2 + \gamma}\biggr)
\end{equation}
where $C$ indicates the number of classes, $n$ indicates the number of class samples, $A_{in}$ indicates those vectors containing all positive examples predicted by the model, and $B_{in}$ indicates the vectors containing all positive examples of the ground truth in the dataset. For the purposes of smoothing, it is common to add a $\gamma$ factor to both the nominator and denominator and in general, $\gamma$ = 1 \cite{dice_loss}. No data augmentation was used in the training phase. We do not apply any preprocessing methods.
\begin{figure*}[t!]
	\centering
		\includegraphics[width=18cm]{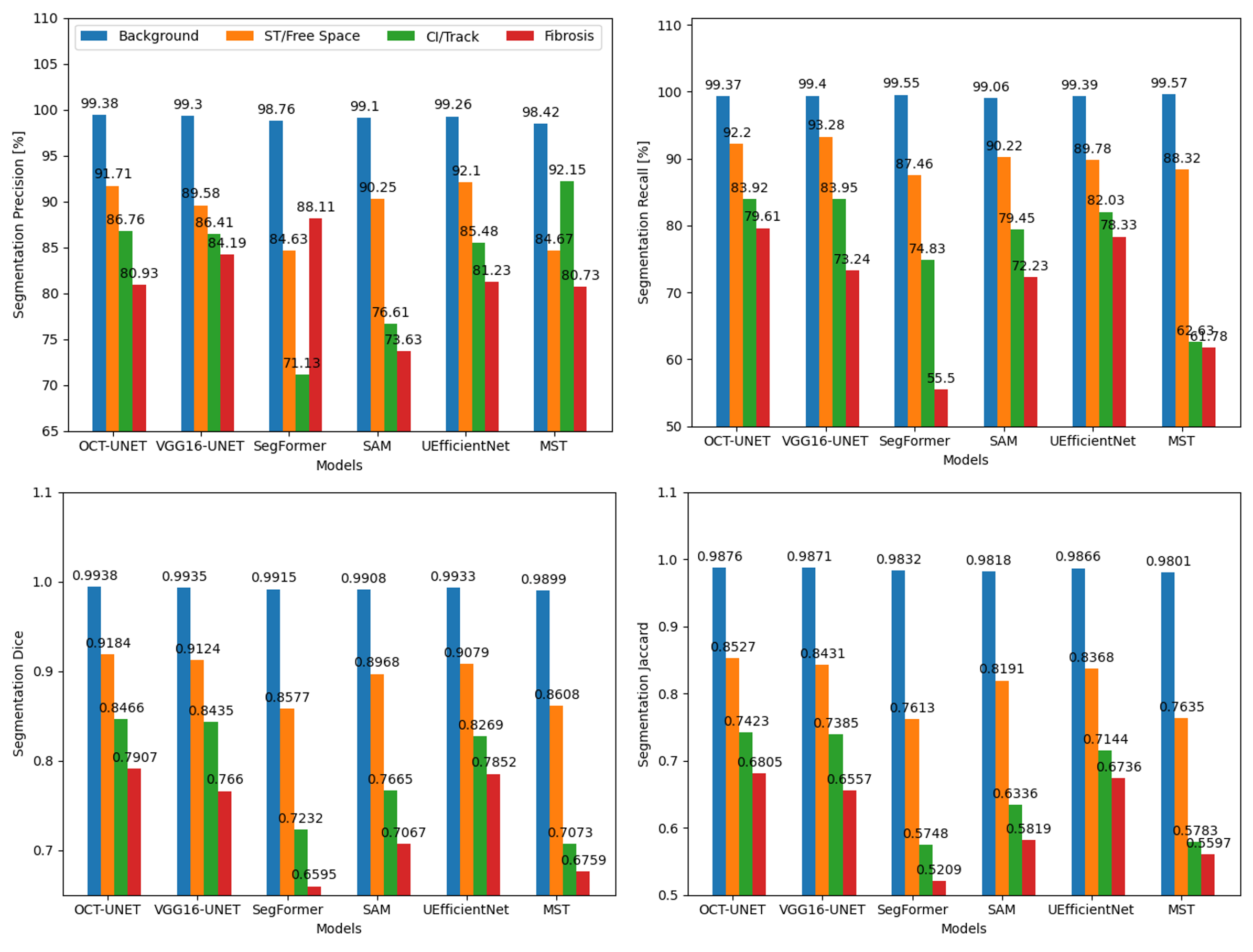}
	  \caption{Quantitative results from benchmarking different models are provided for each class. Average quantitative results are given in Table \ref{tab:quantitative}. }\label{fig:quantitative_barcharts}
\end{figure*}

\begin{table*}[h]
\begin{center}
\caption{Quantitative results averaged over all four classes. Models are ranked according to the DSC metric. Our own 2D-OCT-UNET model outperforms other state-of-the-art semantic segmentation models}
\label{tab:quantitative}
\begin{tabular}{l c c c c c c c}
\hline
Model & Accuracy & Precision & Recall & Dice (DSC) & Jaccard & parameters & model size\\ 
\hline
Proposed 2D-OCT-UNET & \textbf{99.09}\% & 89.70\% & \textbf{88.78}\% & \textbf{0.8874} & \textbf{0.8158} & 34,623,099 & 132.08MB\\

VGG16-UNET \cite{vgg16_unet} & 99.06\% & \textbf{89.87}\% & 87.47\% & 0.8788 & 0.8061 & 25,862,532 & 98.66MB\\

UEfficientNet \cite{UEfficientNet} & 99.03\% & 89.52\% & 87.38\% & 0.8783 & 0.8029 & 113,890,355 & 434.46MB\\

Segment Anything Model (SAM, with adapters) \cite{kirillov2023segany} & 98.71\% & 84.90\% & 85.24\% &  0.8402 & 0.7541 & 94,008,145 & 358.614MB\\


MST-DeepLabv3+ \cite{MST_DeepLab_nature2024} & 98.57\% & 88.99\% & 78.07\% & 0.8085 & 0.7204 & 6,407,940 & 24.44MB\\
SegFormer \cite{segformer} & 98.60\% & 85.66\% & 79.34\% & 0.8080 & 0.7101 & 3,715,684 & 14.17MB\\
\hline
\end{tabular}
\end{center}
\end{table*}

We fine-tune the SegFormer model for 200 epochs using an Adam optimizer with the default learning rate of lr = 0.001. Loss computation is performed internally by the Keras model. 
After computing the loss, the model returns a structured dataclass object which is then used to guide the training process. No data augmentation was used in the training phase. We apply channel-wise standardization of the input images and resize to the $512 \times 512$ input resolution. 

We follow \cite{gu2024build} to fine-tune the SAM model in auto-mode (without prompts) and with adapters. We use batch size of 2 for the dataloader and the image size of $1024 \times 1024$. We set the maximal number of epochs to 200. The fine-tuning process terminates (early stopping) after 78 epochs on our system.



\subsection{Experimental results}
In this section, we present our empirical results of benchmarking the six aforementioned models on the newly collected Cochlear OCT dataset. In order to obtain quantitative results we employ standard objective semantic segmentation metrics such as Accuracy, Precision, Recall, Dice Score Coefficient (DSC) and the Jaccard index also known as the Intersection over Union (IoU). 
\begin{figure*}[t!]
	\centering
		\includegraphics[width=17cm]{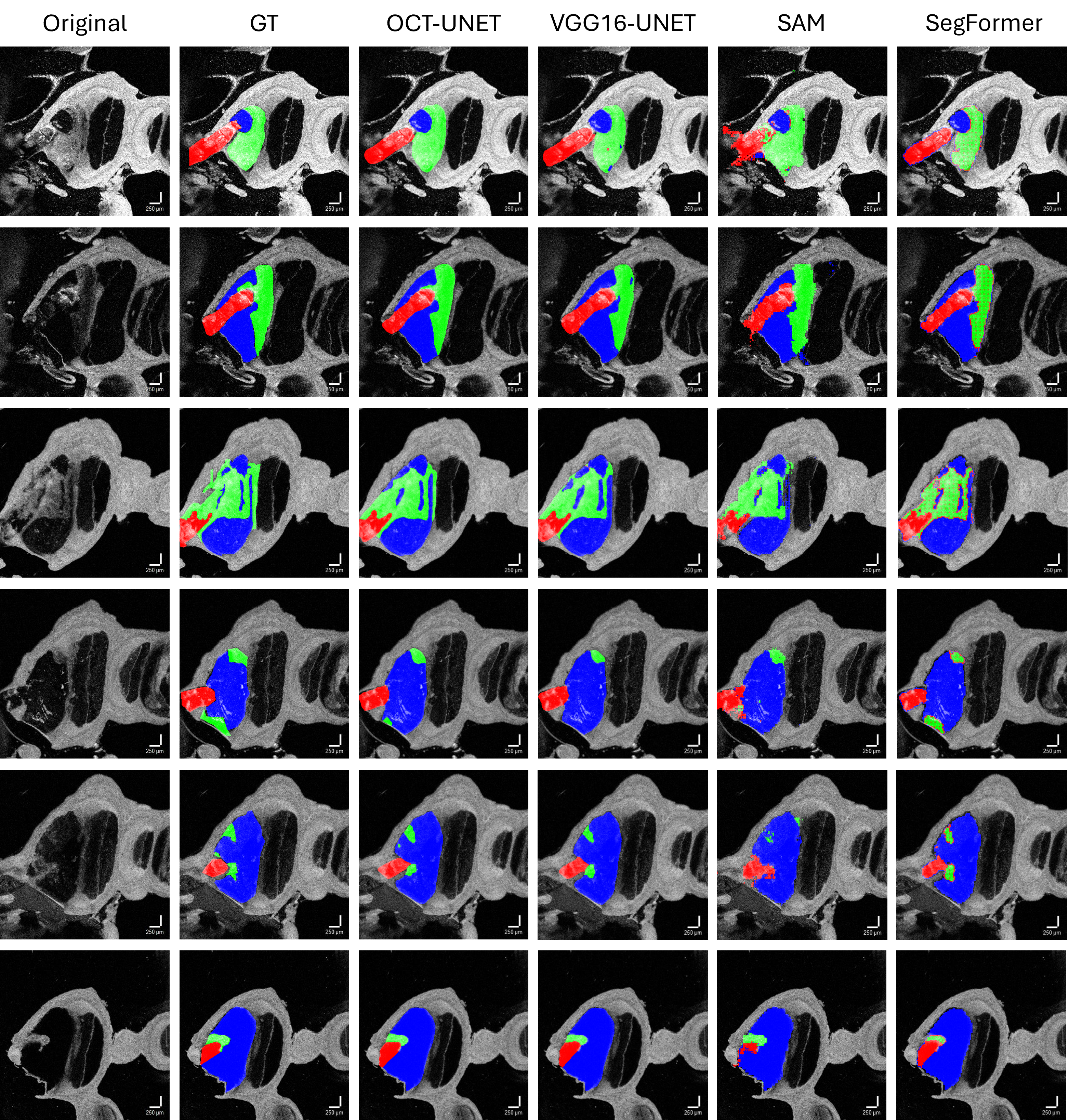}
	  \caption{Samples from qualitative results are provided for each volume. First two rows: OCTV1L, Third row: OCTV7L, Fourth row: OCTV9L, Fifth row: OCTV10L and the Sixth row: OCTV11L. By comparing to the Ground Truth (GT) it can be inferred that the 2D-OCT-UNET model performed the best and the SAM (prompt-free auto-mode implementation with adapters \cite{gu2024build}) model qualitatively performed the worst. As previously, the CI/Track class is depicted in red, the Fibrosis class in green and the ST/Free Space class in blue.}\label{fig:qualitative_results}
\end{figure*}
\begin{figure*}[t!]
	\centering
		\includegraphics[width=17cm]{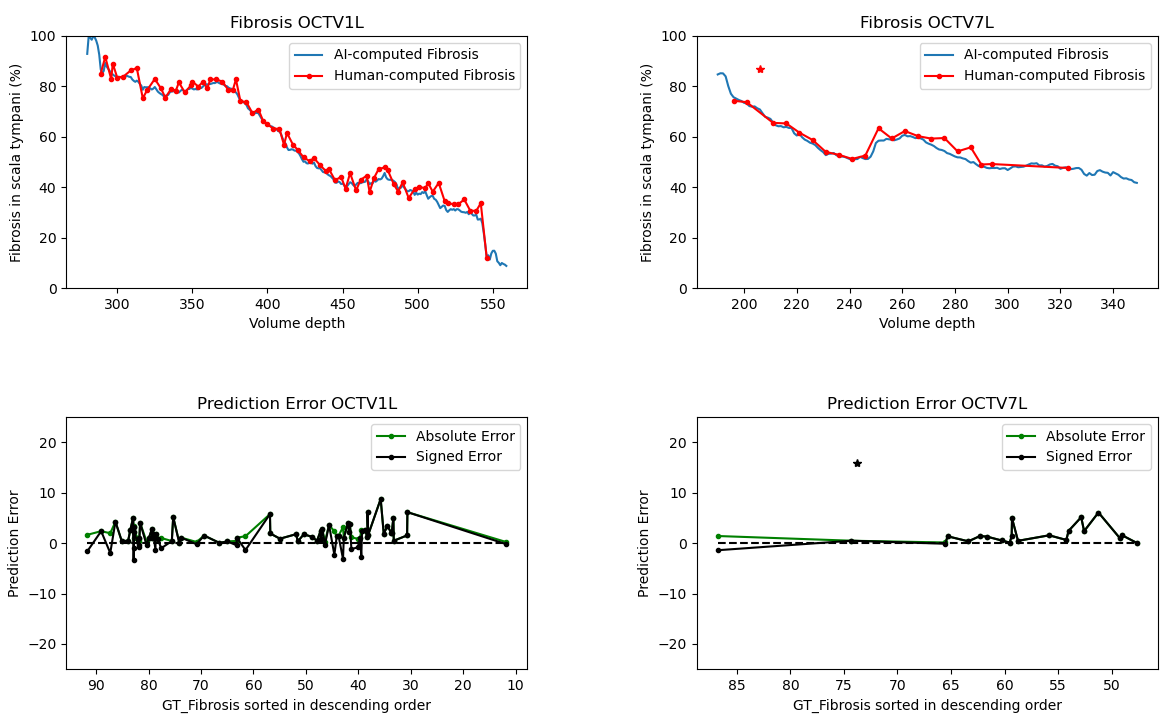}
	  \caption{Example of fibrosis quantification using our 2D-OCT-UNET model. First row shows outcome of human (red curve) and AI-computed (blue curve) segmentation of fibrosis for OCTV1L and OCTV7L volumes. Fibrotic burden is expressed as a function of Z position within the OCT scan. The AI-computed segmentation was trained upon the more sparse human-computed segmentation and applied to all sections from the top of the ST to the point where depth within the preparation reduced Signal-to-Noise-Ratio (SNR) and therefore confidence in segmentation by the human segmenters. The AI performance was excellent and provided a high resolution assessment of the amount of fibrosis in the ST. The second row shows quantitative prediction error analysis plots. Prediction error is defined as a difference between the human-computed (GT) fibrosis and the AI-computed fibrosis. The correlation statistics for the absolute error are as follows: OCTV1L $\rho=-0.1088$, $p=0.3464$, OCTV7L $\rho=0.0374$, $p=0.8723$ where $\rho$ is the Pearson's correlation coefficient and $p$ is the two-sided p-value. The correlation statistics for the signed error are as follows: OCTV1L $\rho=-0.0857$, $p=0.4585$, OCTV7L $\rho=-0.1343$, $p=0.5616$. We identified an outlier in the ground truth of the OCTV7L volume. We do not include this outlier in our formal analysis and plot this datapoint (marked with an asterisk) separately from the error lines. The annotations for the outlier slice are shown in Fig. \ref{fig:outlier_annotations}.}\label{fig:fibrosis_quantification}
\end{figure*}
All these metrics are based on confusion matrix computation in terms of true positives (TP), true negatives (TN), false positives (FP) and false negatives (FN). Accuracy and Precision are defined as follows:
\begin{equation}
    Accuracy=\frac{TP+TN}{TP+TN+FP+FN}
\end{equation}
\begin{equation}
    Precision=\frac{TP}{TP+FP}
\end{equation}
Accuracy represents correctly classified pixels and Precision represents the quality of correct detections relative to the ground truth. Specifically, Recall is the number of true positives divided by the sum of true positives and false negatives. Therefore, high Recall corresponds to a low false negative rate. Dice is a harmonic mean of Precision and Recall of a prediction. In addition, Dice penalizes false positives which are common in highly class-imbalanced medical imaging datasets. Jaccard index is a statistic that measures the IoU of the ground truth and the predicted segmentation. Specifically, it measures how well the model can separate foreground objects from their background. While Jaccard measures the degree of overlap between segmentation masks, Dice computes the similarity between masks penalizing over- and under-segmentation less than Jaccard \cite{metrics}.

Quantitative experimental results are provided in Table~\ref{tab:quantitative} and Fig.~\ref{fig:quantitative_barcharts}.
Qualitative results are shown in Fig.~\ref{fig:qualitative_results}. It can be seen from Table \ref{tab:quantitative} that the proposed 2D-OCT-UNET outperforms other models in four out of five metrics. Addition of adapters to SAM somewhat improves its performance but is still below the performance of the 2D-OCT-UNET.

Interestingly, the SegFormer model quantitatively underperforms the other state-of-the-art semantic segmentation models on our Cochlear OCT dataset by a large margin. Based on the DSC metric, the difference to the 2D-OCT-UNET is 0.0794. This can also be seen in the qualitative plots in Fig.~\ref{fig:qualitative_results} that show the pixelated ``edge'' artefacts of the segmentation maps returned by the SegFormer model. 

We think that this result may be due to the inability of SegFormer to handle the highly imbalanced Cochlear OCT dataset with an unequal distribution of foreground and background pixels in all slices. While we are using the region-based multiclass Dice and the categorical crossentropy losses with other CNN-based models, we hypothesize that the internal loss computation of SegFormer is more prone to the persisting class imbalance.%

Despite the fact that the research community is still far from the complete understanding of the imbalanced semantic segmentation \cite{Neural_Collapse}, there is already a multitude of advanced loss functions available (e.g. Unified Focal Loss \cite{Unified_Focal_Loss}) that address the context of class imbalance in medical imaging and semantic segmentation in CNNs.  Incorporating these loss functions into our training procedure could potentially further improve our current performances. However, this investigation will be reserved for the future work. 

To showcase the application of the best performing 2D-OCT-UNET, we compute the cochlear fibrotic burden (amount of fibrosis in percent that is present in each imaged volume) by using the following equation:
\begin{equation}
    Amount\ of Fibrosis = \frac{A_{Fibrosis} \times 100}{A_{ST\ Free\ Space}+A_{Fibrosis}}
\end{equation}
where $A_{Fibrosis}$ and $A_{ST\_Free\_Space}$ are areas (in pixels) of segmented ``Fibrosis'' and ``ST/Free Space'' classes. 

Fig. \ref{fig:fibrosis_quantification} shows an example of computation for two volumes OCTV1L and OCTV7L where the comparison is provided between the amount of fibrosis computed using the 2D-OCT-UNET model and more sparse computation using manually segmented areas. The high accuracy computation of the AI-computed fibrosis (blue curves) is the direct result of our successful semantic segmentation procedure. 

Fig. \ref{fig:fibrosis_quantification} (first row) shows variation in the best model's prediction accuracy, with lower accuracy for OCTV1L than OCTV7L. The corresponding prediction error plots are shown in the second row of Fig. \ref{fig:fibrosis_quantification}. Prediction error is defined as the difference between the ground truth and the model output fibrosis values. Signed and absolute error are shown, where signed error indicates where the model overestimates (negative values) or underestimates (positive values) the amount of fibrosis relative to ground truth. 

Calculating absolute error gave us the opportunity to assess whether general inaccuracy was attributable to the amount of fibrosis class present in the image. However, we did not find any significant correlation between amount of fibrosis and absolute error (OCTV1L $\rho=-0.1088$, $p=0.3464$, OCTV7L $\rho=0.0374$, $p=0.8723$, where $\rho$ is Pearson's correlation coefficient and $p$ is the $p$-value).  

When signed error was assessed, a small but not significant negative correlation was observed between that and amount of fibrosis for OCTV7L (OCTV1L $\rho=-0.0857$, $p=0.4585$, OCTV7L $\rho=-0.1343$, $p=0.5616$, where $\rho$ is Pearson's correlation coefficient). This finding can be interpreted as a tendency for the model to underestimate fibrotic burden as it reduces in percentage volume of the ST, relative to human segmenters. Coupled with the differing overall error between the models,  we  conclude that this source of model inaccuracy was likely due to ground truth noise. 

\begin{figure}[h]
	\centering
		\includegraphics[width=8cm]{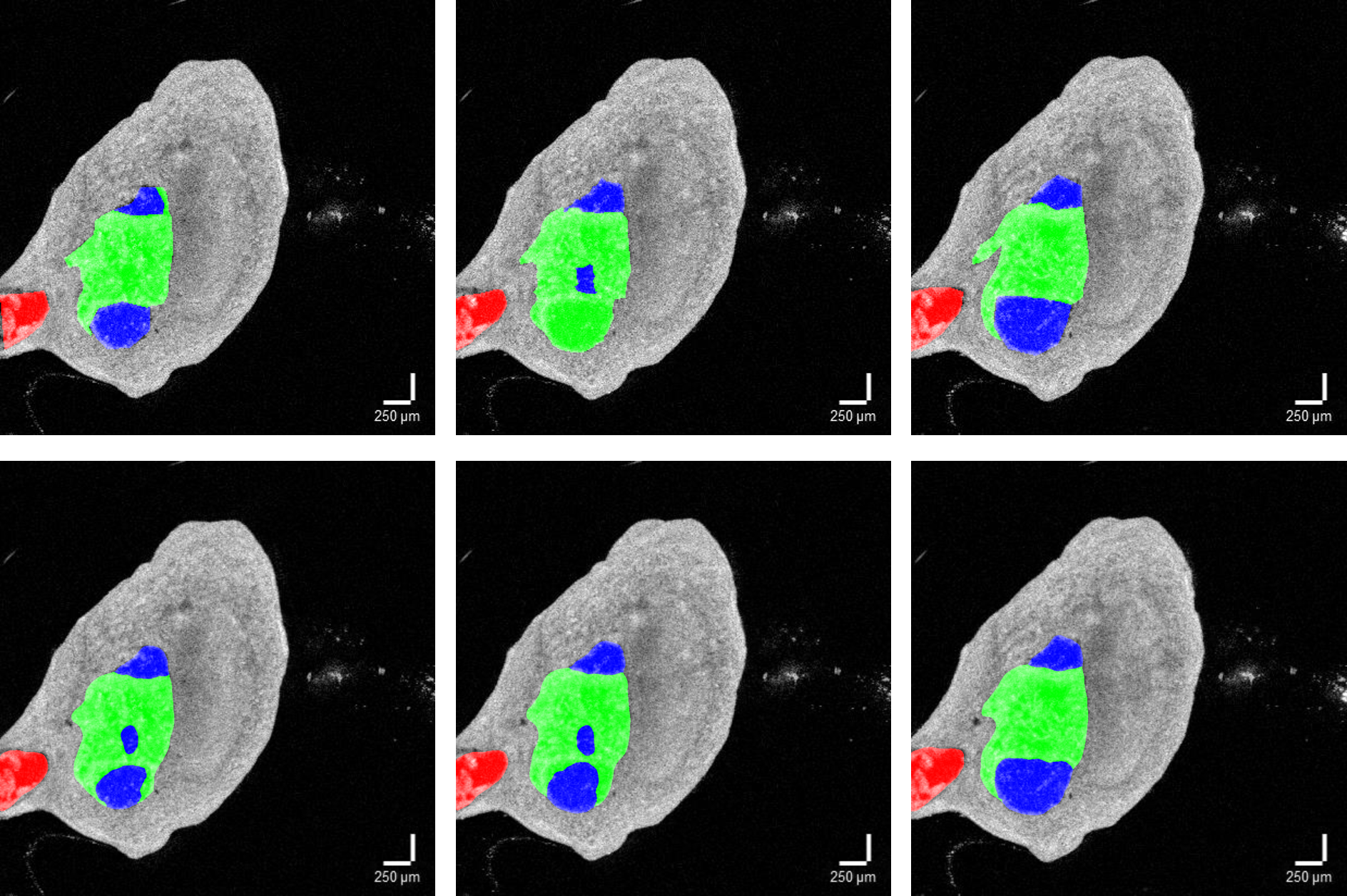}
	  \caption{First row depicts ground truth and the second row depicts corresponding 2D-OCT-UNET predictions for the OCTV7L volume. An outlier in the ground truth annotations can be seen in the middle slice in the first row. This human mistake in the annotation would cause a significant $p$-value of 0.0050 if included in the formal error analysis. }\label{fig:outlier_annotations} 
\end{figure}



\section{Discussion}\label{sec: Discussion}
We developed and applied the best performing model to the cochlear fibrotic burden computation, which is of particular importance for understanding the residual hearing loss pathway in small animal models and its translation to human anatomy. 

A challenge was the possibility of noisy labels, a well-known issue in deep learning based image segmentation \cite{li2022learning}, \cite{noisy_labels_MIA2020}. Instances of noisy labeling cropped up where the basilar membrane-fibrosis interface was difficult to identify. Despite instances of noisy labeling, the model often appeared to correctly identify this interface and not label existing tissue as fibrosis. Future risk of noisy labeling can be mitigated by more segmenter experience with the imaging volume or can be avoided by averaging two or more human measurements. Further, noisy labeling offers an opportunity to train our 2D-OCT-UNET to more actively control for the issue. 

We observed that the fibrotic burden computation is less accurate for the OCTV1L volume. We hypothesize that this performance of the 2D-OCT-UNET on the OCTV1L volume is due to the presence of more noisy labels in that volume and the more imbalanced nature of the semantic class annotations. Fig.~\ref{fig:fibrosis_quantification} (second row) shows quantitative prediction error analysis. It can be observed that there is more uncertainty (more noise) in human measurement with intermediate levels of fibrosis, whereas there is more certainty (less noise) with low or high levels of fibrosis. The intermediate slices in these volumes have more complex morphology, and therefore visual complexity. Certainly, the outlier highlighted in Fig. \ref{fig:outlier_annotations} was tolerated and ignored by 2D-OCT-UNET, implying that our approach can be robust to certain levels of annotation noise. Whether this outcome means that 2D-OCT-UNET can detect extraneous or unusual features within our dataset remains to be tested. 

2D-OCT-UNET was designed with a specific use case in mind: to analyze fibrotic burden in OCT scans of guinea pig cochleae. Currently, our dataset is small at 5 animals, and the efficacy of 2D-OCT-UNET is limited to within-volume training. However, we felt that 2D-OCT-UNET's performance was worth sharing to other CI and computer vision researchers. We have presented segmentation efficacy as a proof of concept, on the so-far analyzed (at the time of writing), relatively small imaging dataset, while more data are being collected and analyzed. Thus, in our future work we will expand the size of the Cochlear OCT dataset, and pursue generalization of 2D-OCT-UNET to working across volumes. We will also anticipate annotating each image by two different segmenters and then averaging the annotations to minimize ground truth error.

It is unclear but compelling to consider how 2D-OCT-UNET will perform on imaging derived from other species, including humans. Increased anatomical variability - not to mention the presence of the CI as a means to treat an underlying pathology and not as a controlled research tool, complicate this adaptation. With these caveats in mind, we speculate that 2D-OCT-UNET may be applicable to other imaging modalities, such as micro computed tomography, which has superior resolution and penetration depth to standard OCT. It would be of great interest to test 2D-OCT-UNET's utility for human temporal bone imaging.

\section{Conclusion}\label{sec: Conclusion}
Our three main contributions are: 
\begin{itemize}
    \item We curate a first-of-its-kind annotated Cochlear OCT dataset from chronically cochlear implanted guinea pigs exhibiting intracochlear fibrosis. 
    \item We introduce a new task of cochlear semantic segmentation to study and quantify cochlear fibrotic burden.
    \item We design a 2D-OCT-UNET model and show its superior performance on the cochlear fibrosis quantification task.
\end{itemize}
The aim of this work is to evaluate the performances of different state-of-the-art deep learning models on the new task of cochlear semantic segmentation on the newly collected Cochlear OCT dataset with CI-induced intracochlear fibrosis present. 

This task is challenging owing to the high variability in shapes, sizes and appearances of the three object classes of interest. All these properties naturally result in the extremely imbalanced dataset. Additionally, 2D-OCT-UNET was only trained on implanted cochleae - training the model on volumes without fibrosis would introduce greater dataset imbalance, but is a necessary next step in model optimization.

We have benchmarked six different CNN-based and vision Transformer-based state-of-the-art models including the foundation Segment Anything Model (SAM). Despite task complexity, the best performing model was our own 2D-OCT-UNET configuration which achieved multiclass DSC of 0.8874. Qualitative results verify the high quality of semantic segmentation. 

Which mechanisms underlie residual hearing loss after cochlear implantation remains undiscovered and several numerical and mathematical models have been proposed \cite{RHL}, \cite{RHL_2016}. We hope that our findings could advance future studies on exploring the relationship between cochlear fibrosis and residual hearing loss. 

\section*{Acknowledgements}\label{sec: Acknowledgements}
This work was supported by funding from Hearing Health Foundation Emerging Research Grant 969672 (Burwood). Research reported in this publication was supported by the National Institute On Deafness And Other Communication Disorders of the National Institutes of Health under Award Number R21DC020794 (Burwood). The content is solely the responsibility of the authors and does not necessarily represent the official views of the National Institutes of Health.

Special thanks to the OHSU medical students Teyhana Rounsavill, Madeline Otto and Adrienne Fettig for their work on the dataset annotation. We would like to acknowledge Prof. Alfred L. Nuttall from  OHSU for his feedback on the manuscript. Special thanks to the two anonymous reviewers for their constructive feedback. Thanks to Cochlear Inc. for vending of preclinical CI arrays. 

\section*{Compliance with ethical guidelines}\label{sec: Ethical}
All procedures were undertaken in accordance with OHSU IACUC protocols and with the permission of the local ethics committee. Protocol numbers: IP00000453 (approved 4/6/2022) and IP00001278 (Approved 12/29/2020, renewed 12/06/2023).





\bibliographystyle{IEEEtran}
\bibliography{casrefs}





\end{document}